\documentclass{article}

\usepackage{times}
\usepackage{graphicx} % more modern
\usepackage{natbib}

\usepackage{algorithm}
\usepackage{algorithmic}
\usepackage[compatibility=false]{caption}%http://52.209.67.144/papercheck.html
\usepackage{algorithm,algorithmic}
\usepackage{amsmath,mathtools,amsfonts}
\usepackage{array}
\usepackage{adjustbox}
\usepackage{graphicx}
\usepackage{verbatim}
\usepackage{float}
\usepackage{subfig}
\usepackage{nicefrac}
\usepackage{enumitem}

\newcommand{\INDSTATE}[1][1]{\STATE\hspace{#1}}

\DeclarePairedDelimiter{\norm}{\lVert}{\rVert}

\newlength{\myheightA}
\newlength{\myheightTopImg}
\usepackage{hyperref}

\usepackage[accepted]{icml2017} 

\icmltitlerunning{Accelerating Eulerian Fluid Simulation With Convolutional Networks}

\usepackage{caption}
\begin{document} 

\twocolumn[
\icmltitle{Accelerating Eulerian Fluid Simulation With Convolutional Networks}



% you can specify symbols, otherwise they are numbered in order
\icmlsetsymbol{equal}{*}

\begin{icmlauthorlist}
\icmlauthor{Jonathan Tompson}{brain}
\icmlauthor{Kristofer Schlachter}{nyu}
\icmlauthor{Pablo Sprechmann}{nyu,deepmind}
\icmlauthor{Ken Perlin}{nyu}
\end{icmlauthorlist}

\icmlaffiliation{nyu}{New York University, New York, USA}
\icmlaffiliation{deepmind}{Google Deepmind, London, UK}
\icmlaffiliation{brain}{Google Brain, Mountain View, USA}

\icmlcorrespondingauthor{Jonathan Tompson}{tompson@google.com}

\icmlkeywords{Naiver-Stokes, fluid, physics, machine learning, ICML}

\vskip 0.3in

% This is left here (commented), on the off chance we have room for it.
% \begin{center}
%     \centering
% \adjustbox{height=\myheightTopImg}
%       {\includegraphics[width=\textwidth]{figures/plume.png}} \hfill
% \adjustbox{height=\myheightTopImg}
%       {\includegraphics[width=\textwidth]{figures/arc.png}} \hfill
% \adjustbox{height=\myheightTopImg}
%       {\includegraphics[width=\textwidth]{figures/bunny.png}} \hfill
%     \captionof{figure}{Smoke simulation using our system. Videos can be found in the supplemental materials, or at: \\
%     \protect\url{http://cims.nyu.edu/~schlacht/CNNFluids.htm}}
%     \label{fig:pics}
% \end{center}%
]



\printAffiliationsAndNotice{}  % leave blank if no need to mention equal contribution
%\footnotetext{hi}

% Pablo suggestions:
% Yang model to "lightweight".
% Learning curves
% Figure 8 and 9 the same thing.
% Real Yang in Figure 8.
% Sub-heading in results for comparison with Yann.

% Model figures to 1 figure.
% Leave table.
% 

% **********************************************************************
\begin{abstract} 
% **********************************************************************
Efficient simulation of the Navier-Stokes equations for fluid flow is a long standing problem in applied mathematics, for which state-of-the-art methods require large compute resources. In this work, we propose a data-driven approach that leverages the approximation power of deep-learning with the precision of standard solvers to obtain fast and highly realistic simulations. Our method solves the incompressible Euler equations using the standard operator splitting method, in which a large sparse linear system with many free parameters must be solved. We use a Convolutional Network with a highly tailored architecture, trained using a novel unsupervised learning framework to solve the linear system. We present real-time 2D and 3D simulations that outperform recently proposed data-driven methods; the obtained results are realistic and show good generalization properties.
\end{abstract}

% **********************************************************************
\section{Introduction}
% **********************************************************************
\label{sec:introduction}

Real-time simulation of fluid flow is a long standing problem in many application domains: from computational fluid dynamics for industrial applications, to smoke and fluid effects for computer graphics and animation. 
% In particular, physically-accurate graphics simulation has achieved remarkable results animating fluids like water or smoke. 
% However, there is a big gap between pre-computing a simulation for several hours on high-end computers and simulating it in real-time with limited compute resources. 
High computational complexity of existing solutions has meant that real-time simulations have been possible under restricted conditions. In this work we propose a data-driven solution to the invicid-Euler equations that is faster than traditional methods, while remaining competitive in long-term simulation stability and accuracy. This work is focused on computer graphics animation as the driving application for our architecture, however the techniques presented here can be extended to more complicated forms of Navier-Stokes (with appropriate modifications that are outside the scope of this work).
% While several works have studied different alternatives for mitigating this limitation, computational complexity remains one of the most important constraints.

The dynamics of a large number of physical phenomenon are governed by the incompressible Navier-Stokes equations, which are a set of partial differential equations that must hold throughout a fluid velocity field for all time steps. There are two main computational approaches for simulating these equations: the Lagrangian methods, that approximate continuous quantities using discrete moving particles~\cite{gingold1977smoothed}, and the Eulerian, methods that approximate quantities on a fixed 
% regular % Jie Tan comment: some grids are not regular (i.e. octree).
grid~\cite{FosterRealisticAnimation}. We adopt the latter for this work.

%We adopt the standard operator splitting method to solve the inviscid Euler equations. 
Eulerian methods are able to produce accurate results simulating fluids like water with high compute costs. The most demanding portion of this method is the ``pressure projection" step, which satisfies an incompressibility constraint. It involves solving the discrete Poisson equation and leads to a well-known sparse, symmetric and positive-definite linear system. Exact solutions can be found via traditional convex optimization techniques, such as the Preconditioned Conjugate Gradient (PCG) algorithm or via stationary iterative methods, like the Jacobi or Gauss-Seidel methods. PCG exhibits fast asymptotic convergence, however it suffers from high time-constants and is generally not suited to GPU hardware. The Jacobi method is used for real-time applications, however it suffers from poor asymptotic convergence. Additionally, the computational complexity of both these algorithms is strongly \emph{data-dependent} (i.e. boundary condition dependent) and these iterative methods are truncated in real-time to fit within a computational budget.
%Many efficient numerical methods have been proposed to mitigate this limitation (we discuss these approaches in Section \ref{sec:relatedwork}), however real-time simulation on high resolution domains remains infeasible.
%As such, Jacobi or Gauss-Seidel methods are used for real-time applications as they are trivially parallelizable, while they suffer from poor asymptotic convergence and are terminated using a fixed number of iterations. In this regime, these ``exact-solvers'' are treated as approximate methods since few guarantees can derived for the magnitude of the linear system residual (i.e. velocity divergence) given a fixed number of iterations. Additionally, the computational complexity of both these algorithms is strongly \emph{data-dependent}.
%The most common Lagrangian approach for real-time simulation is the Smoothed Particle Hydrodynamics \cite{gingold1977smoothed} (SPH) method that approximates continuous quantities using discrete particles. 
%It is a grid-free method in which the coordinates move with the fluid (particles).
%The Eulerian approach that we adopt for this work, approximates quantities on a regular grid.

In this paper, we propose a machine learning based approach for accelerating this linear projection that is fast and whose complexity is \emph{data-independent}. We leverage the power of deep-learning techniques to derive an approximate linear projection. We claim that our machine-learning approach can take advantage of the statistics of fluid data and the local sparsity structure of the discrete Poisson equation to learn an approximate method that ensures long-term stability using a fixed computational budget. The main contributions of this work are as follows:
% Convolutional Networks (ConvNets) are powerful machines particularly well suited for modeling the local correlations of image data. ConvNets are used in most modern computer vision systems, achieving state-of-the-art results in many fundamental problems~\cite{Inceptionv4}.
%
\vspace{-1mm}
\vspace{-\topsep}
\begin{enumerate}[label=(\roman*)]
\setlength{\parskip}{0pt}
\setlength{\itemsep}{0pt plus 1pt}
\item We rephrase the learning task as an unsupervised learning problem; since ground-truth data is not required we can incorporate loss information from multiple time-steps and perform various forms of non-trivial data-augmentation. 
\item We propose a collection of domain-specific ConvNet architectural optimizations motivated by the linear system structure itself, which lead to both qualitative and quantitative improvements. We incorporate an in-line normalization scheme, we devise a multi-resolution architecture to better simulate long range physical phenomena and we formulate the high level solver architecture to include a ``pressure bottleneck'' to prevent trivial solutions of our unsupervised objective.
\item Our proposed simulator is stable and is fast enough to permit real-time simulation. Empirical measurements suggest good generalization properties to unseen settings.
\item We provide a public dataset and processing pipeline to procedurally generate random ground-truth fluid frames for evaluation of %data-driven 
simulation methods.
\end{enumerate}
\vspace{-\topsep}
\vspace{-1mm}

%Note that, while we do not need ground-truth label information, our model does benefit from an efficient sampling of realistic initial conditions. 
%That is, the space of all divergent velocity is unconstrained, and so our network's generalization performance is improved when using a dataset of ``realistic'' initial conditions that approximately samples the manifold of real-world fluid simulation states. 

%In this regime, these ``exact-solvers'' are treated as approximate methods since few guarantees can derived for the magnitude of the linear system residual (i.e. velocity divergence) given a fixed number of iterations. Additionally, the computational complexity of both these algorithms is strongly \emph{data-dependent}.

The paper is organized as follows. We discuss related work in Section~\ref{sec:relatedwork}. In Section~\ref{sec:fluidequations} we briefly introduce the fluid simulation techniques used in this paper. In Section~\ref{sec:model} we present the proposed model. We provide implementation details in Section~\ref{sec:datasettraining}.
Experimental results are described in Section~\ref{sec:results} and conclusion are drawn in Section~\ref{sec:conclusion}.

% **********************************************************************
\section{Related Work}
% **********************************************************************
\label{sec:relatedwork}

Recent work has addressed the problem of computational performance in fluid simulations. \cite{mcadams2010parallel} proposed a multi-grid approach as a preprocessing step of the PCG method. This method can significantly improve performance in large scene settings. However, multi-grid methods are very difficult to implement, hard to parallelize on the GPU, have non-trivial ``failure-cases'' (e.g. multi-grid complexity resorts to single-resolution complexity for simulations with highly irregular domain boundaries), and the method still requires a iterative optimization procedure with data-dependent complexity.

Some methods propose inexact (but efficient) approximate solutions to the Poisson equation, such as iterated orthogonal projections \cite{molemaker2008low} or coarse-Grid corrections \cite{lentine2010novel}. While these approaches can be competitive in low resolution settings, they are data agnostic: they do not exploit the statistics of the data to be processed.
A natural approach is to tackle the problem in a data-driven manner - by adapting the solver to the statistics of the data of interest. 
%
%A family of works aims at accelerating the solution of linear system solve 
The main idea of data-driven methods is to reduce computation by operating on a representation of the simulation space of significantly lower dimensionality. 
%These representations are obtained by exploiting the structure (or redundancy) in the data. 
%There is a natural compromise between computational cost and the richness of the representation.
%
The Galerkin projection transforms the dynamics of the fluid simulation to operations on linear combinations of pre-processed snap-shots \cite{TreuilleModelReduction,DeWittLaplacianEignenfunctions}.
In \cite{RaveendranBlendingLiquids} the authors propose generating complex fluid simulations by interpolating a relatively small number of existing pre-processed simulations.
The state graph formulation \cite{StantonStateRefining} leverages the observation that on simple simulations only a small number of states are visited. 

More recently - and most related to this work - some authors have regarded the fluid simulation process as a supervised regression problem. These approaches train black-box machine learning systems to predict the output produced by an exact solver using random regression forests \cite{DataDrivenFluidSimulation} or neural networks \cite{yang2016data} for Lagrangian and Eulerian methods respectively. Ladick\'{y} et al., propose an adaptation of regression forests for smoothed particle hydrodynamics. Given a set of hand-crafted features corresponding to individual forces,
%and the incompressibility constraint of the Navier-Stokes equations,
the authors trained a regressor for predicting the state of each particle in the following time-step. Yang et al. train a patch-based neural network to predict the ground truth pressure given local previous-frame pressure, voxel occupancy, and velocity divergence of an input frame.

A major limitation of existing learning-based methods is that they require a dataset of linear system solutions provided by an exact solver. Hence, targets cannot be computed during training and models are trained to predict the ground-truth output always starting from an initial frame produced by an exact solver, while at test time this initial frame is actually generated by the model itself. This discrepancy between training and simulation can yield errors that can accumulate quickly along the generated sequence.
This problem is analogous to that encountered when generating sequences with recurrent neural networks, see \cite{bengio2015scheduled} and references therein. Additionally, the ConvNet architecture proposed by Yang et al. is not suited to our more general use-case; in particular it cannot accurately simulate long-range phenomena, such as gravity or buoyancy. While providing encouraging results that offer a significant speedup over their PCG baseline, their work is limited to data closely matching the training conditions (as we will discuss in Section~\ref{sec:results}).

%
%In the context of Lagrangian methods, a position based fluid (PBF) approach has been proposed in \cite{macklin2013position}. In this method, all particles are first advected and then projected to satisfy the incompressibility condition. Although this requires an iterative procedure, PBF allows the use of larger time-steps than previous alternatives. To further speed-up the computation, a multi-scale approach was introduced in \cite{horvath2013mass}.
%
%In \cite{ChentanezLargeBodies} the authors propose a grid-based solution that can produce efficient simulations when restricted to simpler topology. 
Finally we point out that our work is related to the emergent field of artificial intelligence (AI) referred as intuitive or naive physics. Despite the tremendous progress achieved in AI, agents are still far from achieving common sense reasoning, which is thought to play a crucial role in the development of general AI. Recent works have attempted to build neural models that can make predictions about stability, collisions, forces and velocities from images or videos, or interactions with an environment (e.g. \cite{lerer2016learning}). In this context, having accurate and efficient models for simulating a real world environment could be crucial for developing new learning systems \cite{hamrickimagination,fleuret2016predicting,byravan2016se3}. We believe that our work could be used in this context in the more challenging setting of an agent interacting with fluids, see for instance \cite{kubricht2016probabilistic}.

%\pablo{Any ML references? Physics ICML.}

% **********************************************************************
\section{Fluid Equations}
% **********************************************************************
\label{sec:fluidequations}

When a fluid has zero viscosity and is incompressible it is called \textit{inviscid}, and can be modeled by the Euler equations~\cite{batchelor1967introduction}:
\begin{eqnarray}
\frac{\partial u}{\partial t} =  -u \cdot \nabla u - \frac{1}{\rho}\nabla p + f \label{eq:momentum} \\
\nabla \cdot u = 0 \label{eq:divfree}
\end{eqnarray}
Where $u$ is the velocity (a 2D or 3D vector field), $t$ is time, $p$ is the pressure (a scalar field), $f$ is the summation of external forces applied to the fluid body (buoyancy, gravity, etc) and $\rho$ is fluid density. 
%This formulation has been used extensively in computer graphics applications as it can model a large variety of smoke and fluid effects.
% Gradient and Laplacian operators are computed spatially (in 2D or 3D). 
Equation~\ref{eq:momentum} is known as the momentum equation, it arrises from applying Newton's second law to fluid motion and describes how the fluid accelerates given the forces acting on it.  Equation~\ref{eq:divfree} is the incompressibility condition, which enforces the volume of the fluid to remain constant throughout the simulation.
%Equation~\ref{eq:divfree} is the incompressibility condition and Equation~\ref{eq:momentum} is known as the momentum equation, and we will focus on this version of the Navier-Stokes equations for this work. 
For readers unfamiliar with fluid mechanics and it's associated prerequisite topics (multi-variable calculus, finite-difference methods, etc), we highly recommend ~\cite{bridson2008fluid} as an introductory reference to this material.

We numerically compute all spatial partial derivatives using finite difference (FD) methods on a MAC grid~\cite{mac_grid_harlow}. The MAC grid representation samples velocity components on the face of voxel cells, and the scalar quantities (e.g. pressure or density) at the voxel center. This representation removes the non-trivial nullspace of central differencing on the standard uniformly sampled grid and it simplifies boundary condition handling, since solid-cell boundaries are collocated with the velocity samples.

Equations~\ref{eq:momentum} and \ref{eq:divfree} can be solved via the standard operator splitting method, which involves an advection update and a ``pressure projection" step. Here is an overview of the single time-step velocity update algorithm:

\begin{algorithm}
  \caption{Euler Equation Velocity Update}
  \label{alg:the_alg}
  \begin{algorithmic}[1]
%    \STATE Choose a time-step $\Delta t$ \label{alg:time}
    \STATE Advection and Force Update to calculate $u_t^\star$: \label{alg:advection}
      \INDSTATE[0.2cm] (optional) Advect scalar components through $u_{t-1}$ \label{alg:scalaradvect}
      \INDSTATE[0.2cm] Self-advect velocity field $u_{t-1}$ \label{alg:veladvect}
      \INDSTATE[0.2cm] Add external forces $f_{body}$
      \INDSTATE[0.2cm] Add vorticity confinement force $f_{vc}$
      \INDSTATE[0.2cm] Set normal component of solid-cell velocities.
    \STATE Pressure Projection to calculate $u_t$: \label{alg:project}
      \INDSTATE[0.2cm] Solve Poisson eqn, $\nabla^2 p_t=\frac{1}{\Delta t}\nabla \cdot u_t^\star$, to find $p_t$ \label{alg:pressure}
      \INDSTATE[0.2cm] Apply velocity update $u_t = u_{t-1} - \frac{1}{\rho}\nabla p_t$
  \end{algorithmic}
\end{algorithm}

At a high level, Algorithm~\ref{alg:the_alg} step~\ref{alg:advection} ignores the pressure term ($-\nabla p$ of Equation~\ref{eq:momentum}) to create an advected velocity field, $u_t^\star$, which includes unwanted divergence, and then step~\ref{alg:project} solves for pressure, $p$, to satisfy the incompressibility constraint (Equation~\ref{eq:divfree}). This produces a divergence free velocity field, $u_t$. It can be shown that an exact solution of $p$ in step~\ref{alg:pressure}, coupled with a semi-Lagrangian advection routine in steps~\ref{alg:scalaradvect} and \ref{alg:veladvect}, results in an unconditionally stable numerical solution.
In our hybrid apporach we modify step~\ref{alg:pressure} by replacing the exact projection for a \emph{learned} one.
% Therefore, $\Delta t$ in step~\ref{alg:time} is somewhat arbitrary, and often heuristics are used to minimize visual artifacts. 

For advection of scalar fields and self-advection of velocity, we perform a semi-Lagrangian backward particle trace using the Maccormack method~\cite{maccormack}. When the backward trace would otherwise sample the input field inside non-fluid cells (or outside the simulation domain), we instead clamp each line trace to the edge of the fluid boundary and sample the field at its surface.

% Notably absent from the Euler Momentum Equation~\ref{eq:momentum} is the Navier-Stokes term for viscosity ($\nu \nabla \cdot \nabla u$ - where $\nu$ is the kinematic viscosity). We choose to ignore this term since our Semi-Lagrangian advection scheme, as well as our ConvNet formulation for solving pressure, introduces unwanted numerical dissipation which acts as viscosity.\footnote{Note that our method does not prohibit the use of viscosity, rather that we find it unnecessary for our application. One can trivially extend Algorithm \ref{alg:the_alg} to include a viscosity component after the velocity advection step \ref{alg:veladvect}.} 
We use vorticity confinement~\cite{steinhoff1994modification} to counteract unwanted numerical dissipation, which attempts to reintroduce small-scale detail by detecting the location of vortices in the flow field and then introducing artificial force terms to increase rotational motion around these vortices. This firstly involves calculating the vorticity strength, $w = \nabla \times u$, using central difference and then calculating the per-voxel force, $f_{vc} = \lambda h \left(N \times w\right)$, 
% $f_{vc}$:
%\begin{equation*}
%f_{vc} = \lambda h \left(N \times w\right)
%\end{equation*}
%
where, $N = \nicefrac{\nabla\left|w\right|}{\norm{\nabla\left|w\right|}}$, $\lambda$ controls the amplitude of vorticity confinement, and $h$ is the grid size (typically $h=1$).

Algorithm~\ref{alg:the_alg} step~\ref{alg:pressure} is by far the most computationally demanding component. It involves solving the following Poisson equation:
\begin{equation}
\nabla^2 p_t=\frac{1}{\Delta t}\nabla \cdot u_t^\star
\label{eq:poisson}
\end{equation}

Rewriting the above equation results in a large sparse linear system $A p_t = b$, where $A$ is referred to in the literature as the 5 or 7 point Laplacian matrix (for 2D and 3D grids respectively).
% Typically the Preconditioned Conjugate Gradient (PCG) method is used to solve the above linear system.
Despite $A$ being symmetric and positive semi-definite, the linear system often has a large number of free parameters, which means that with standard iterative solvers a large number of iterations must be performed to produce an adequately small residual. Furthermore, this number of iterations is strongly data-dependent. In this paper, we use an alternative machine learning (and data-driven) approach to solving this linear system, where we train a ConvNet model to infer $p_t$. The details of this model will be covered in Section~\ref{sec:model}.
 
After solving for pressure, the divergence free velocity is calculated by subtracting the FD gradient of pressure, $u_t = u_t^\star - \frac{1}{\rho}\nabla p_t$.

To satisfy slip-condition boundaries at fluid to solid-cell interfaces, we set the velocity of MAC cells so that the component along the normal of the boundary face is equal to the normal component of the object velocity (i.e. $\hat{n} \cdot u_{\text{fluid}} = \hat{n} \cdot u_{\text{solid}}$). The MAC grid representation makes this trivial, as each solid cell boundary is at the sampling location of the velocity grid.

% **********************************************************************
\section{Pressure Model}
% **********************************************************************
\label{sec:model}

% The main contribution of this paper is the re-formulation of Algorithm~\ref{alg:the_alg} step~\ref{alg:pressure} (pressure projection step) as a data-driven regression problem in which the regression function is given by a tailored (and highly non-linear) ConvNet. In this section, we describe the proposed architecture and training procedure.
% Exactly satisfying the incompressibility condition of the Euler Equations, $\nabla \cdot u = 0$, using a small and fixed amount of compute per frame has been a long standing goal in not only the computer graphics community, but in the wider applied mathematics community as well. 
%
In traditional formulations, incompressibility is obtained only when the pressure is a solution of the linear system of equations given in Equation~\ref{eq:poisson}.
However, in real-time applications, PCG or Jacobi iterations are truncated before reaching convergence. Therefore the obtained velocity fields are divergent, which may lead to bad solutions (i.e. volume change of smoke or fluid and other visual artifacts) and even instability\footnote{Advection should only be done in a divergence-free velocity field and typically there are no long-term (or multi-frame) mechanisms to ensure divergence is not accumulated.}.
As such, truncation forgoes treating the incompressibility condition as a hard constraint even if this is not the intention, and few guarantees can be given for quality of the divergence residual at termination.
This is especially true in degenerate cases when the matrix in the sparse linear system $A$ has a large number of free-parameters (for example with highly irregular geometry boundaries).
%, which results in the need for a large number of iterations to achieve convergence. 

At the heart of this problem is the ability to predict the run-time of PCG iterations as a function of the required accuracy and the specific data in which it is applied. While there is a vast amount of literature in convex optimization, how data complexity impacts convergence rate of convex solvers is still not well understood~\cite{oymak2015sharp}.
Recent works have attempted to shed some light on these questions \cite{oymak2015sharp,giryes2016tradeoffs}. These results show that, given a fixed computational budget (allowing for only a small number of iterations) in projected gradient approaches, it is worthwhile using very inaccurate projections that may lead to a worse solution in the long-term, but are better to use with the given computational constraints. While this line of work is promising, current results only apply to random systems (such as Gaussian maps) and specific types of input data (with local low dimensional structure) in order to characterize the form of the inaccurate projections for a given problem. 
In the general case, given a fixed computational budget, there is currently no way of guaranteeing a pre-fixed tolerance with respect to the optimality conditions for all possible inputs.

%Alternatively, recent approaches have proposed a data-driven approach in which approximate inference mechanisms are learned from the data itself. In the context of sparse inference, recent approaches have studied the used of deep-learning methods to develop real-time approximate inference mechanisms that profit from the specifics of the data on which they are applied \cite{gregor2010learning,sprechmann2015learning}.
%
The fundamental observation is that, while there is no closed form solution and a numerical solution might be difficult to compute, the function mapping input data to the optimum of an optimization problem is deterministic. Therefore one can attempt to approximate it using a powerful regressor such as deep neural network.
%The obtained results show a very significant improvement when compared to truncating the iterations of the exact solvers.
% In this work we extend these ideas to the case of the Euler equations.
Building upon this observation, another key contribution of this work is that the learning task can be phrased as a completely unsupervised learning problem, if an appropriate ConvNet architecture is used. Instead of using supervised training to directly infer and score the next frame velocity (or pressure) - where the loss would be some distance measure to ground-truth data - we measure the squared L-2 norm of the divergence of the predicted velocity and minimize it directly:
\begin{align}
f_{obj} &= \sum_i w_i \left\{ \nabla \cdot \hat{u}_t\right\}_i^2 \nonumber \\
        &= \sum_iw_i \left\{ \nabla \cdot \left(u_t^\star - \frac{1}{\rho}\nabla \hat{p}_t\right)\right\}_i^2 \label{eq:objfunc}
\end{align}
Where $\hat{u}_t$ and $\hat{p}_t$ are the predicted divergence free velocity and pressure fields respectively and $w_i$ is a per-vertex weighting term which emphasizes the divergence of voxels on geometry boundaries:
\begin{equation*}
w_i=\operatorname{max}\left(1, k - d_i\right)
\end{equation*}
where $d_i$ is a distance field with value 0 for solid cells, and for fluid cells is the minimum Euclidean distance of each fluid-cell to the nearest solid cell (i.e. a signed distance field encoding of the occupancy grid). Since the fluid-solid border represents a small fraction of the domain (due to the sparse nature of the occupancy grid), without importance weighting it contributes a small fraction of the overall loss-function, and might be ignored by our limited capacity ConvNet. However the effect of this term is not significant and has a minor contribution to the simulation quality.

\begin{figure}
    \centerline{\includegraphics[width=\columnwidth]{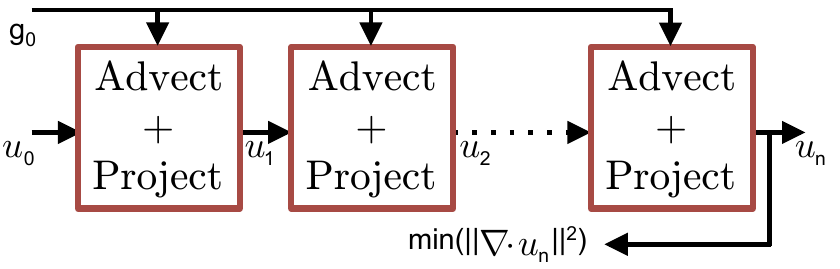}}
    \caption{``Long-term'' Velocity Divergence Minimization}
    \label{fig:future_velocity}
\end{figure}

%We show that the correct pressure (which would otherwise be obtained from supervised labels) uniquely minimizes our unsupervised objective. 
The Equation~\ref{eq:objfunc} formulation has two significant benefits. Firstly, since obtaining ground-truth velocity from expensive iterative methods is no longer necessary, we can perform non-trivial data-augmentation to the input velocity fields. Secondly, we can incorporate loss information from a composition of multiple time-steps without the need of running exact solvers for each frame. With this, we can further improve our ConvNet prediction and long-term stability by adding an additional term to our objective function that minimizes ``long-term'' divergence. We do so by stepping forward the simulation state from $u_0$ for each training sample, to a small number of time-steps $n$ in the future (i.e. we calculate $\hat{u}_n$).\footnote{When training our model we step forward either $n=4$ steps, with probability $0.9$, or $n=25$ with probability $0.1$. Furthermore we use a random time-step - to promote time-step invariance - according to $\Delta t=1/30 * \left(0.203 + \left|N\left(0,1\right)\right|\right)$, where $N\left(0, 1\right)$ is a random sample from a Normal Gaussian distribution.} Then we calculate the scalar value of Equation~\ref{eq:objfunc} using this future frame and we add this to the global objective (to be minimized by SGD). This process is depicted in Figure~\ref{fig:future_velocity}. 

To infer $\hat{p}_t$ we use a Convolutional Network architecture ($f_{conv}$) parameterized by its weights and biases, $c$, and whose input is the divergence of the velocity field, $\nabla \cdot u^\star_t$, and a geometry field $g_{t-1}$ (a Boolean occupancy grid that delineates the cell type for each voxel in our grid: fluid cell or solid cell):
\begin{equation}
\hat{p}_t = f_{conv}\left(c, \nabla \cdot u^\star_t, g_{t-1}\right)
\end{equation}
We then optimize the parameters of this network, $c$, to minimize the objective $f_{obj}$ using standard deep-learning optimization approaches; namely we use Back Propagation (BPROP) to calculate all partial derivatives
%~\cite{lecun1998gradient}
and the ADAM~\cite{kingma2014adam} optimization algorithm to minimize the loss.

\begin{figure}
    \centerline{\includegraphics[width=0.8\columnwidth]{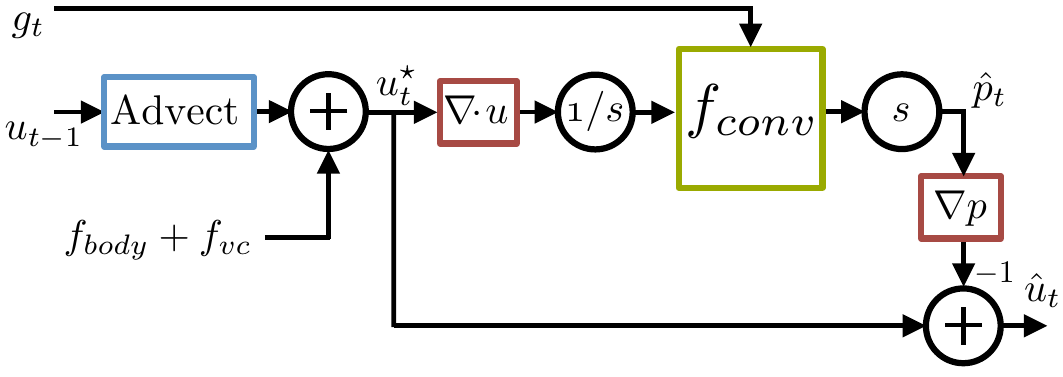}}
    \caption{Velocity-Update Architecture}
    \label{fig:topmodel}
\end{figure}

A block diagram of our high-level model architecture is shown in Figure~\ref{fig:topmodel}, and shows the computational blocks required to calculate the output $\hat{u}_t$ for a single time-step. The \emph{advect} block is a fixed function unit that encompasses the advection step of Algorithm~\ref{alg:the_alg}. After advection, we add the body and vorticity confinement forces. We then calculate the divergence of the velocity field $\nabla \cdot u_t^\star$ which, along with geometry, is fed through a multi-stage ConvNet to produce $\hat{p}_t$. We then calculate the pressure divergence, and subtract it from the divergent velocity to produce $\hat{u}_t$. Note that the bottle-neck architecture avoids obtaining trivial solutions: the perturbation applied by the CNN to the divergent velocity is restricted to be a conservative vector field (i.e. the gradient field with potential $\hat{p}_t$).
Note that the only block with trainable parameters is the ConvNet model and that all blocks are differentiable.

Since the ConvNet ($f_{conv}$) solves a linear system $Ap_t=b$, we are free to scale the left and right hand sides by some constant $s$; we calculate the standard deviation of the input velocity field, $s = \operatorname{STD}\left(u_t^\star\right)$, and normalize the input divergence by this scale value. We then undo the output pressure scale by multiplying by the scale reciprocal. By scale normalizing the input the learned network is made globally scale invariant (which helps generalization performance). Recall that while the network is learning a linear projection, the network itself is highly non-linear and so would otherwise be sensitive to input scale.

%\pablo{Need a comment here about how optimal p uniquely minimizes the objective function, or at least that there's no trivial solution.}

% ConvNets (or deep-networks) are extremely well suited to the task of solving the sparse linear system. In addition, modern GPU architectures are extremely efficient at calculating the ``embarrassingly parallel" convolutional features of deep-networks.

\begin{figure}
\centerline{\includegraphics[width=1.0\columnwidth]{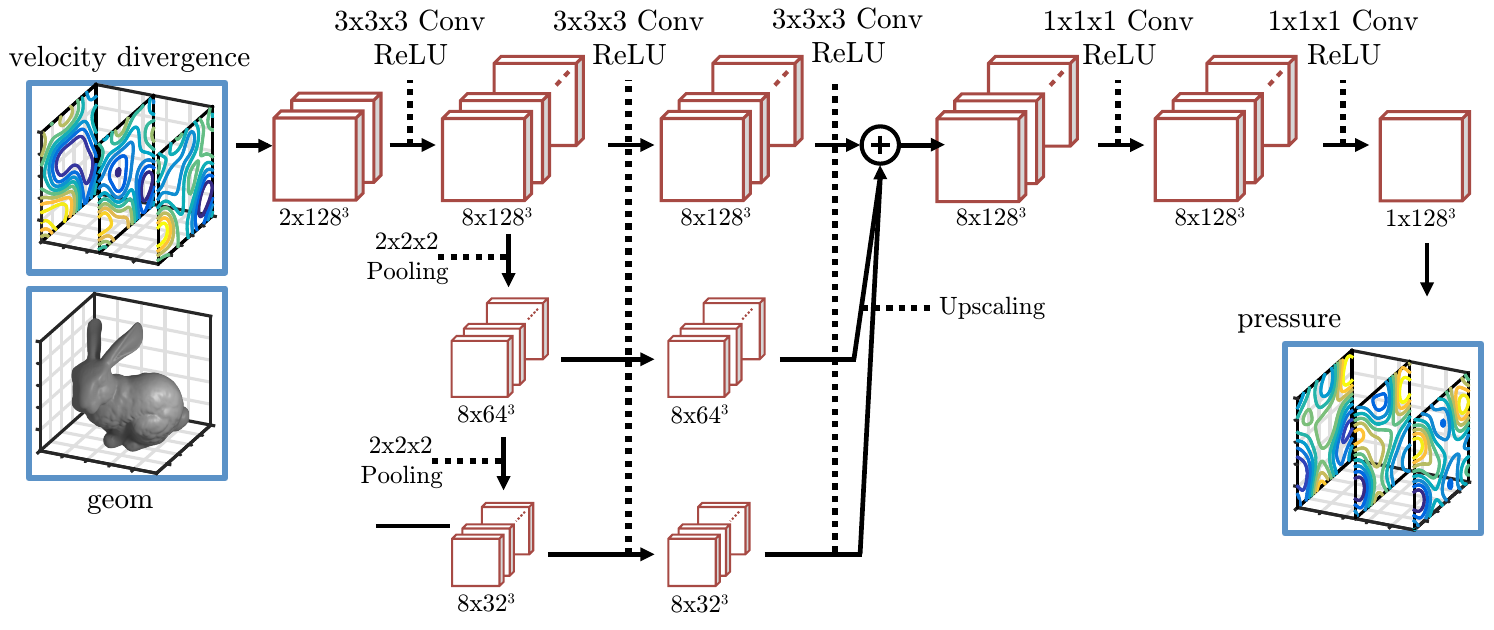}}
\caption{Convolutional Network for Pressure Solve}
\label{fig:convnet}
\end{figure}

The internal structure of the ConvNet architecture is shown in Figure~\ref{fig:convnet}. It consists of 5 stages of convolution (spatial or volumetric for 2D and 3D respectively) and Rectifying Linear layers (ReLU). The convolutional operator itself mimics the local sparsity structure of our linear system; at each layer, features are generated from local interactions and these local interactions have higher-level global behavior in the deeper layers of the network. However a single resolution network would have limited context which limits the network's ability to model long-range external forces (for example the presence of gravity in a closed simulation domain results in a low frequency pressure gradient). As such, we add multi-resolution features to enable modeling long range physical phenomenon by downsampling the first hidden layer twice (average pooling), processing each resolution in parallel then upsampling (bilinear) the resultant low resolution features before accumulating them.

Note that since our network is fully-convolutional, the size of the domain can be modified at inference time; while we train at $64^3$ and $128^2$ resolutions for the 3D and 2D models respectively, the network can perform inference on any size domain.

% Note that we do not claim the network architecture of Figure~\ref{fig:convnet} is in any way ``optimal'' for the task. Its simplicity and relatively small number of trainable parameters allows for fast real-time inference. 
Further improvements can be made to the architecture of Figure~\ref{fig:convnet}. We have experimented with residual connections~\cite{He2015}, gated convolutions~\cite{dauphin2016language} and significantly deeper network architectures. These techniques do improve accuracy but at the cost of added run-time and latency. Alternatively, run-time can be reduced by replacing full-rank convolution kernels with learned separable convolution stages (i.e. compositions of low-rank convolutions) or by using recent model compression techniques~\cite{bengioFewMults,hinton2015distilling} for moderate increases in output divergence.

Why not to use a ConvNet to learn an end-to-end mapping that predicts the velocity field at each time-step? The chaotic change of velocity between frames is highly unstable and easily affected by external forces and other factors. We argue that our proposed hybrid approach restricts the learning task to a stable projection step relieving the need of modeling the well understood advection and external body forces. 
The proposed method takes advantage of the understanding and modeling power of classic approaches, supporting enhancing tools such as vorticity confinement. 
Having said this, end-to-end models are conceptually simpler
and, combined with adversarial training \cite{goodfellow2014generative}, have shown promising results for difficult tasks such as video prediction \cite{mathieu2015deep}. In that setting, fluid simulation is a very challenging test case and our proposed method represents an important baseline in terms of both accuracy and speed.

% **********************************************************************
\section{Dataset Creation and Model Training}
% **********************************************************************
\label{sec:datasettraining}

Note that while we do not need label information to train our ConvNet, our network's generalization performance improves when using a dataset that approximately samples the manifold of real-world fluid states. \textit{To this end, we propose a procedural method to generate a corpus of initial frames for use in training.}

In lieu of real-world fluid data, we use synthetic data generated using an offline 3D solver, mantaflow~\cite{manta}. We then seed this solver with initial condition states generated via a random procedure using a combination of \emph{i}. a pseudo-random turbulent field to initialize the velocity \emph{ii}. a random placement of geometry within this field, and \emph{iii}. procedurally adding localized input perturbations. We will now describe this procedure in detail.

% The difficulty in generating synthetic data for this task is defining a pseudo-random procedure to create training data that, with a reasonable amount of training samples, sufficiently covers the space of input velocity fields. This is clearly a difficult task as this spans all possible $\mathbb{R}^3$ vector fields and geometry. However, despite this seemingly intractable input space, empirically we have found that a plausible model can be learned by defining

Firstly, we use the wavelet turbulent noise of \cite{waveletturbulance} to initialize a pseudo-random, divergence free velocity field. At the beginning of each simulation we randomly sample a set of noise parameters (e.g. wavelet spatial scale and amplitude) and we generate a random seed, which we then use to generate the velocity field.

\begin{figure}
  \centering
    \subfloat{%
      \includegraphics[width=0.32\columnwidth]{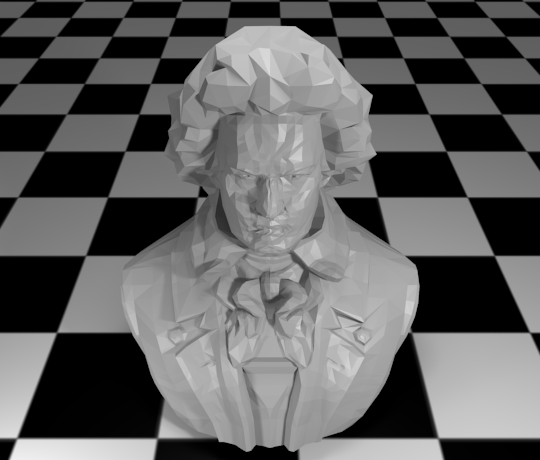}
    }
    \hfill
    \subfloat{%
      \includegraphics[width=0.32\columnwidth]{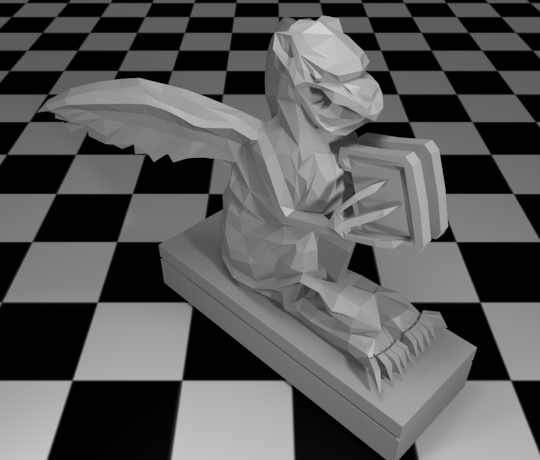}
    }
    \hfill
    \subfloat{%
      \includegraphics[width=0.32\columnwidth]{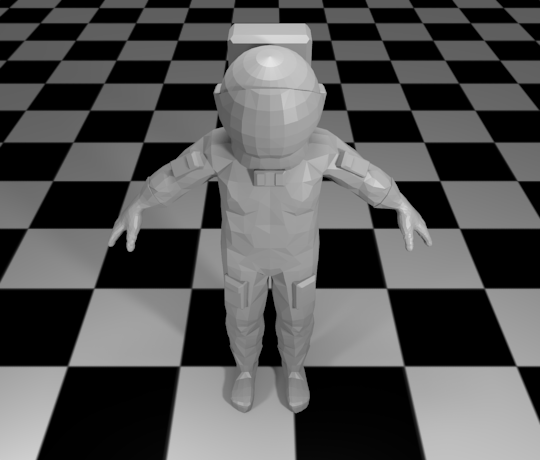}
    }
%     \vspace{-10px}
%     \subfloat{%
%       \includegraphics[width=0.32\columnwidth]{figures/Y10_747.png}
%     }
%     \hfill
%     \subfloat{%
%       \includegraphics[width=0.32\columnwidth]{figures/Y62_elephant.png}
%     }
%     \hfill
%     \subfloat{%
%       \includegraphics[width=0.32\columnwidth]{figures/Y74_Locomotive.png}
%     }
    \caption{A selection of 3D Models used in our dataset}
    \label{fig:3d_models}
  \end{figure}

Next we generate an occupancy grid by selecting objects from a database of models and randomly scaling, rotating and translating these objects in the simulation domain. We use a subset of 100 objects from the NTU 3D Model Database~\cite{3dmodels}; 50 models are used only when generating training set initial conditions and 50 for test samples. Figure~\ref{fig:3d_models} shows a selection of these models. 
%Each model is voxelized using the binvox library~\cite{binvox}.
For generating 2D simulation data, we simply take a 2D slice of the 3D voxel grid. 
Finally, we simulate small divergent input perturbations by modeling inflow moving across the velocity field using a collection of emitter particles of random time duration, position, velocity and size.
%We do this by generating a random set of emitters (with random time duration, position, velocity and size) and adding the output of these emitters to the velocity field throughout the simulation.
% These emitters are procedurally animated to smoothly move around the simulation domain between time-steps. This mimics the sort of divergent inflow one might find due to either user-input or objects moving within the scene.

With the above initial conditions, we use Manta to calculate $u_t^\star$ by advecting the velocity field and adding forces. We also step the simulator forward 256 frames (using Manta's PCG-based solver), recording the velocity every 8 steps.
We generate a training set of 320 ``scenes'' (each with a random initial condition) and a test set of an additional 320 scenes. Each ``scene" contains 32 frames 0.8 seconds apart. We use a disjoint set of geometry for the test and training sets to test generalization performance.
%Since adjacent frames have a large amount of temporal coherence, we can decimate the frame set without loss of ConvNet accuracy, and so we save the synthetic frame to disk every 4 frames. This provides a total of 5120 training set and 5120 test set examples, each with a disjoint set of geometry and with different random initial conditions. 
The dataset is public, as well as the code for generating it.

During training, we further increase dataset coverage by performing data-augmentation. When stepping forward the simulator to calculate long-term divergence, we randomly add gravity, density and vorticity confinement of varying strengths and with a random gravity vector. 

% **********************************************************************
\section{Results and Analysis}
% **********************************************************************
\label{sec:results}

The model of Section~\ref{sec:model} was implemented in Torch7~\cite{torchNips}, with two CUDA baseline methods for comparison; a Jacobi-based iterative solver and a PCG-based solver (with incomplete Cholesky L0 preconditioner). For sparse linear system operations, we used primitives from NVIDIA's cuSPARSE and cuBLAS libraries.

To implement the model of \cite{yang2016data} for comparison, we rephrase their patch-based architecture as an equivalent sliding window model comprised of a 3x3x3 conv and sigmoid stage followed by 3 stages of 1x1x1 convolutions and sigmoid with appropriate feature sizing. Note that this equivalent reimplementation as a ConvNet is significantly faster, as we can make use of the highly optimized convolution implementations from NVIDIA's cudnn library. Yang et al. report 515ms per frame on a 96 x 128 x 96 grid, while our implementation of their model takes only 9.4ms per frame at this resolution.

\begin{figure}
\centering
\subfloat{%
\includegraphics[width=0.32\columnwidth,trim=20cm 0 20cm 8cm, clip=true]{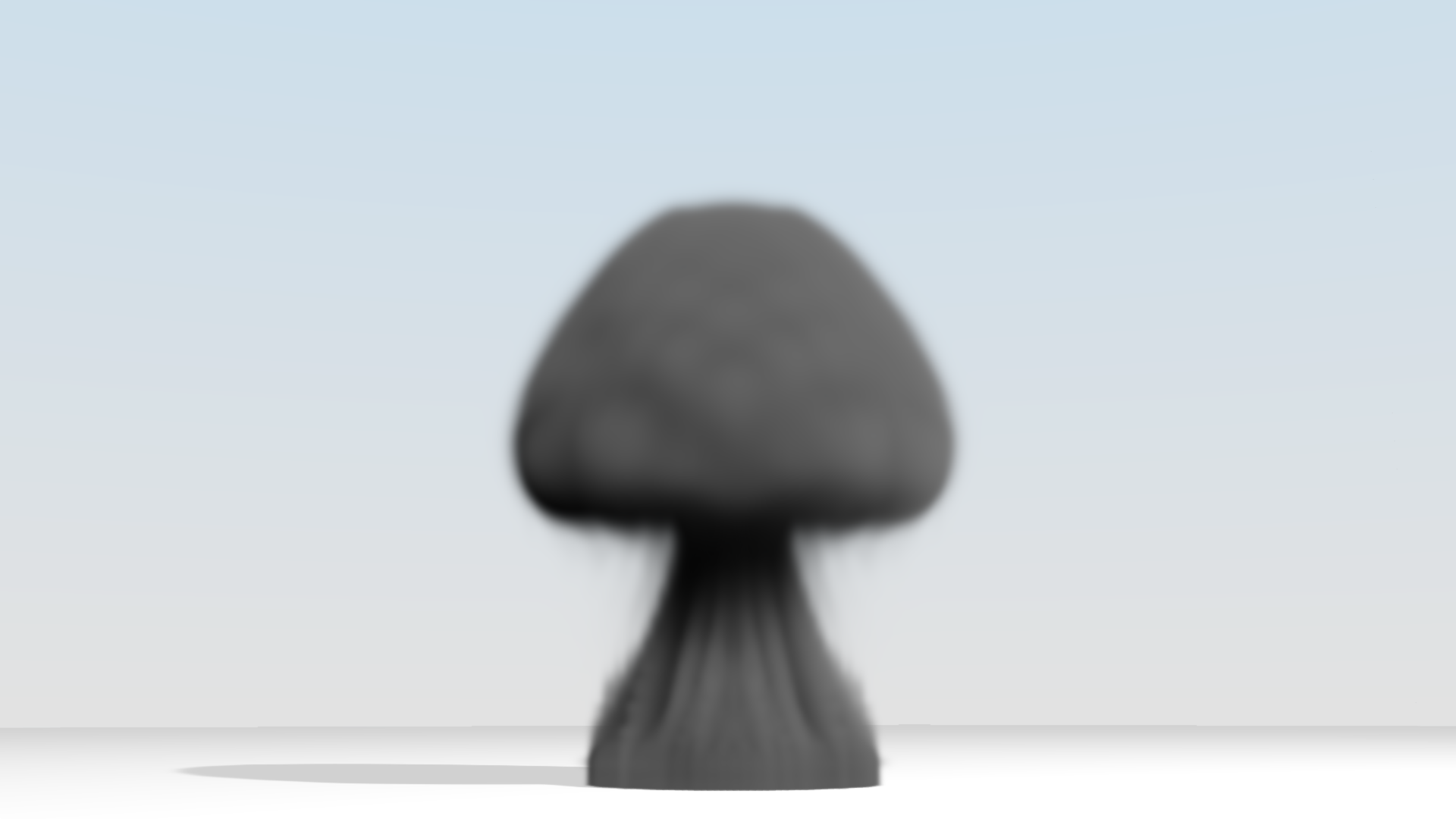}
}
\hfill
\subfloat{%
\includegraphics[width=0.32\columnwidth,trim=20cm 0 20cm 8cm, clip=true]{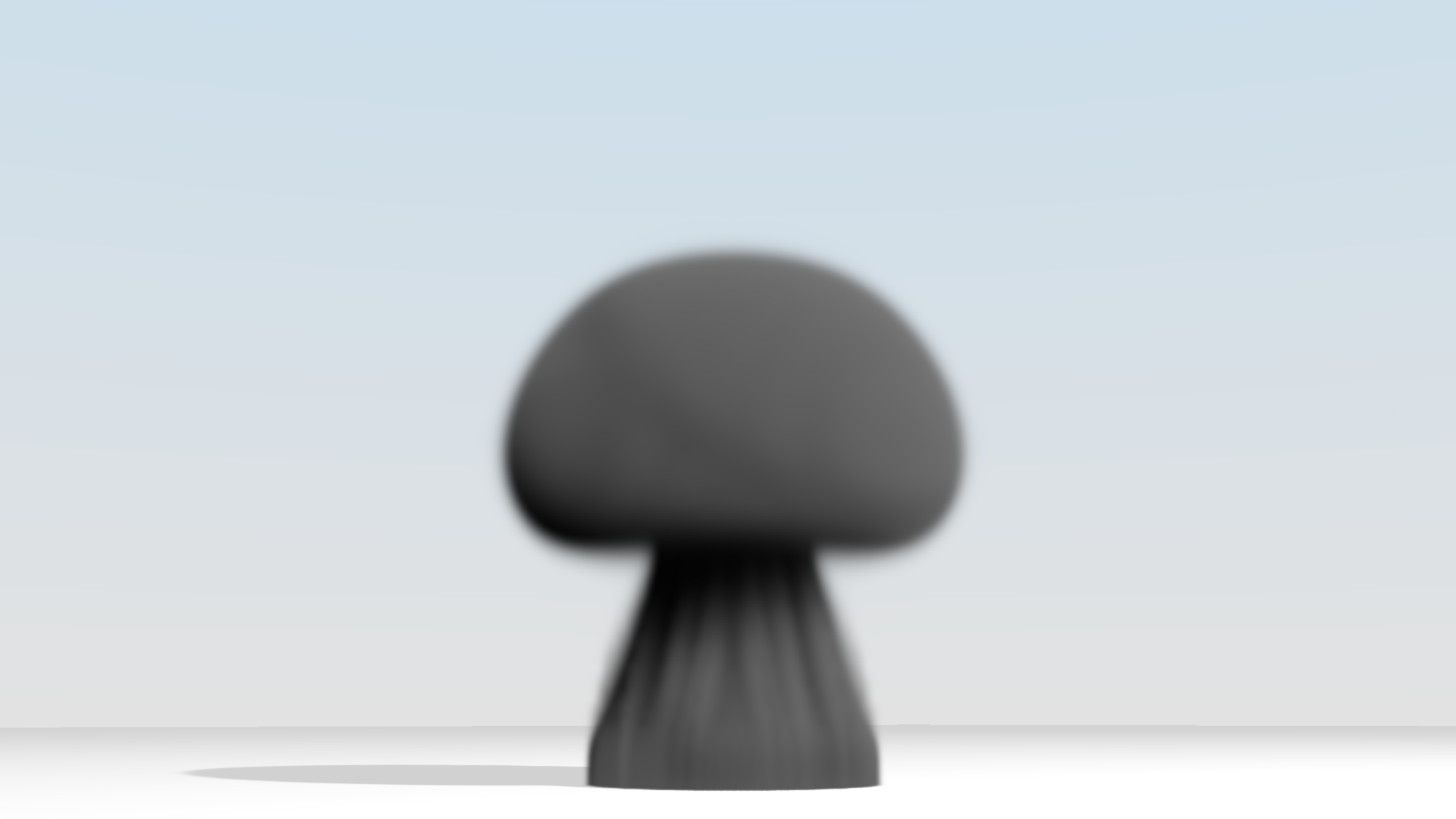}
}
\hfill
\subfloat{%
\includegraphics[width=0.32\columnwidth,trim=20cm 0 20cm 8cm, clip=true]{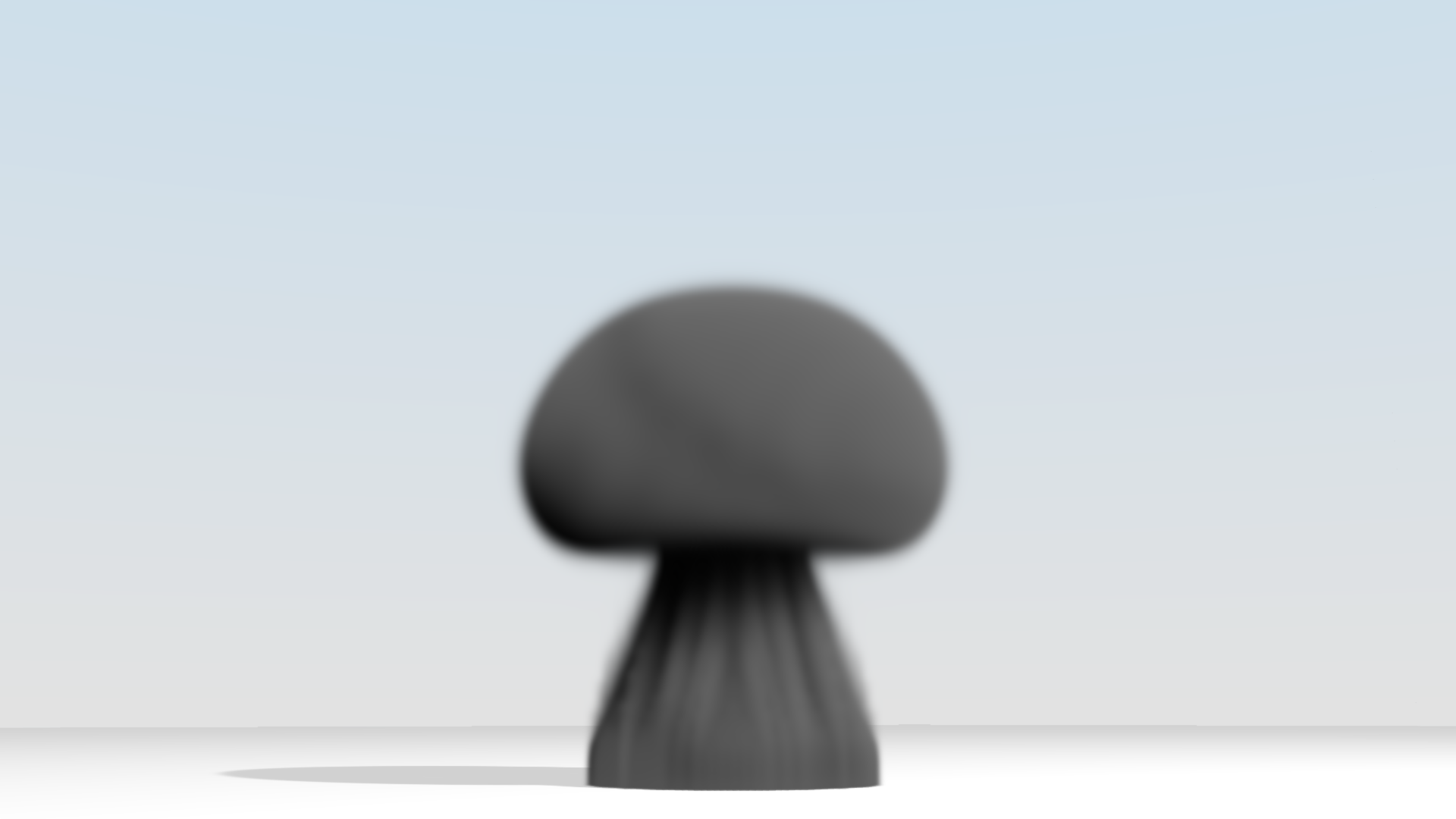}
}

\vspace{-10px}
\subfloat{%
\includegraphics[width=0.32\columnwidth,trim=20cm 0 20cm 8cm, clip=true]{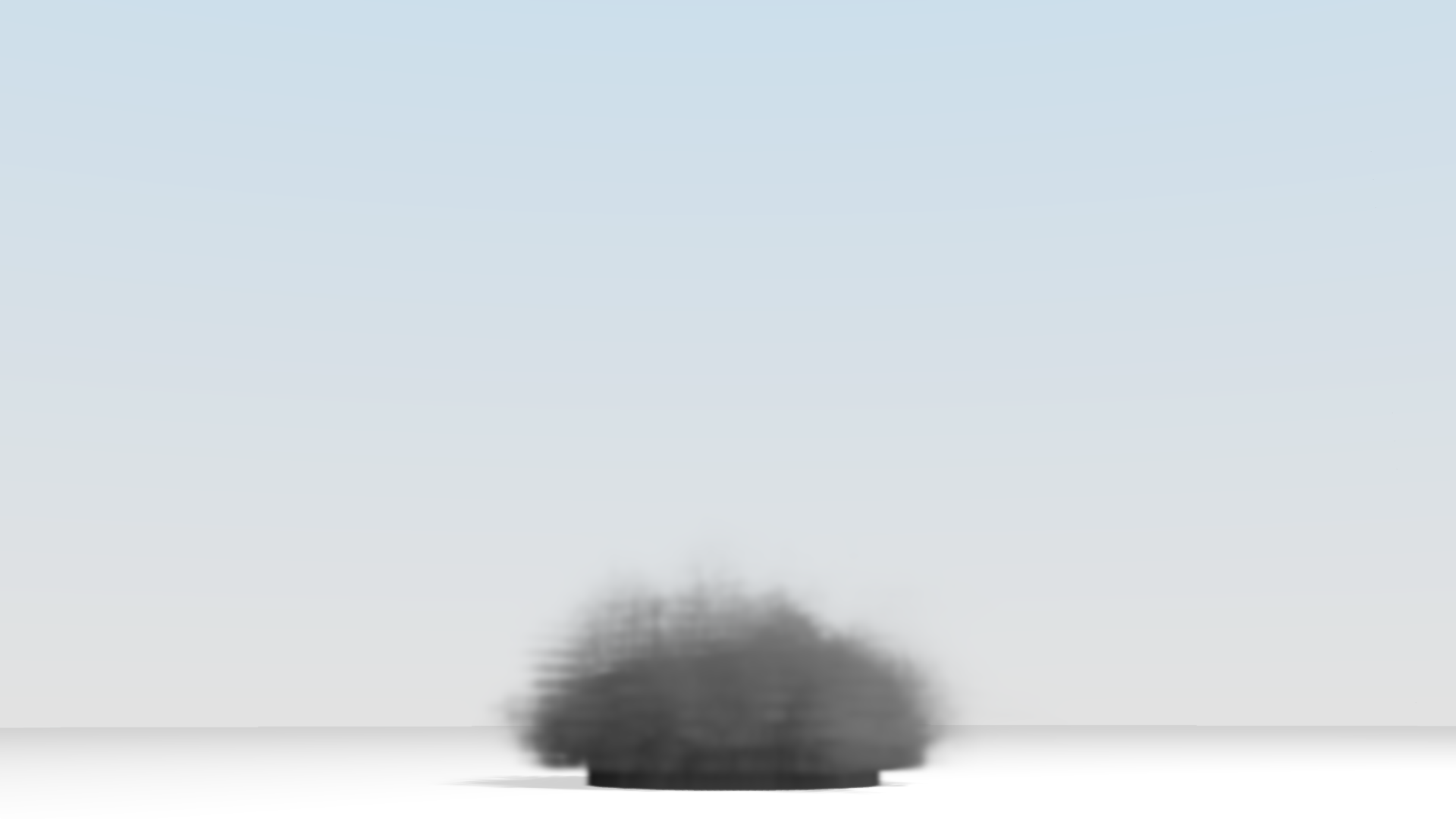}
}
\hfill
\subfloat{%
\includegraphics[width=0.32\columnwidth,trim=20cm 0 20cm 8cm, clip=true]{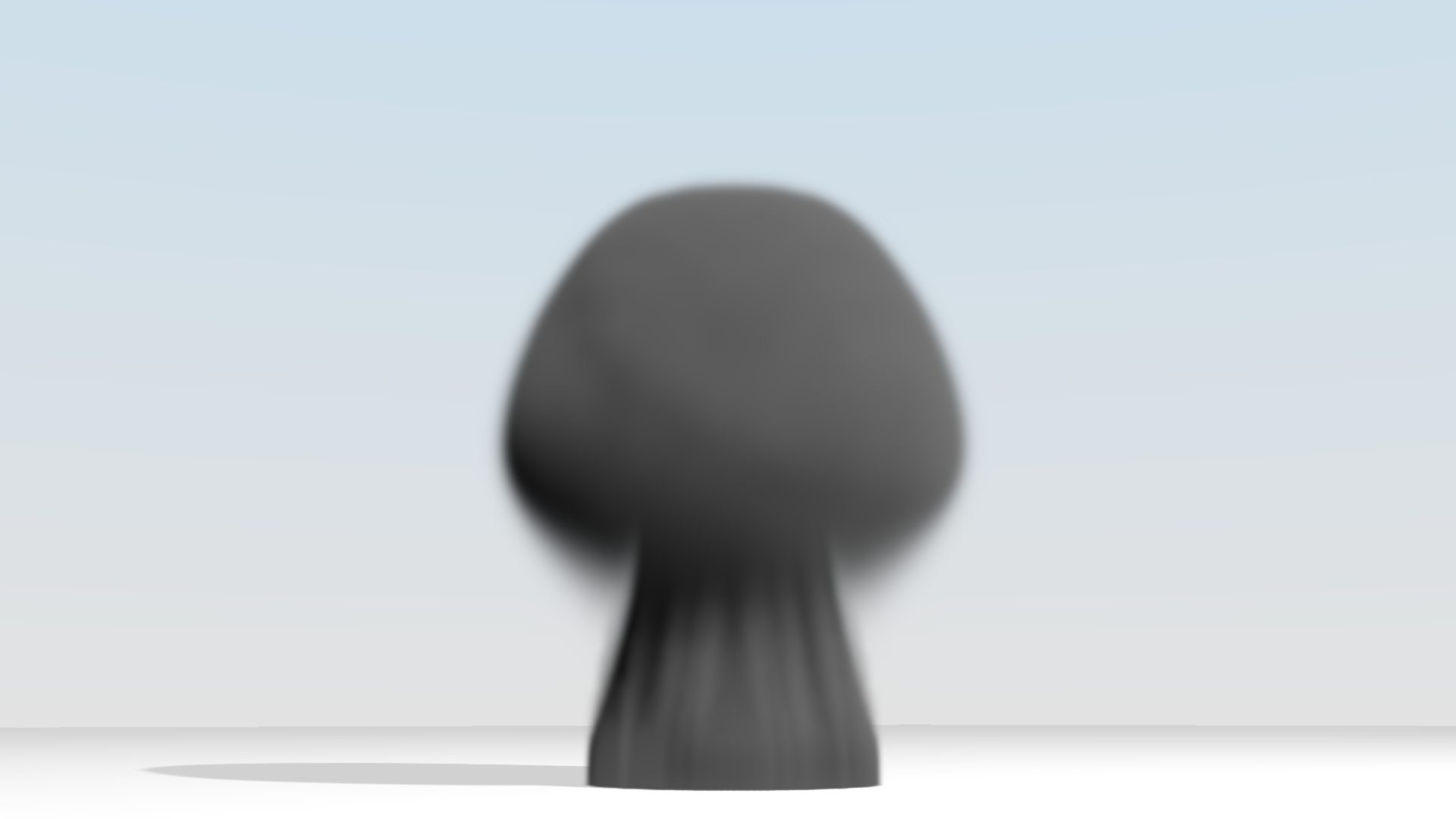}
}
\hfill
\subfloat{%
\includegraphics[width=0.32\columnwidth,trim=20cm 0 20cm 8cm, clip=true]{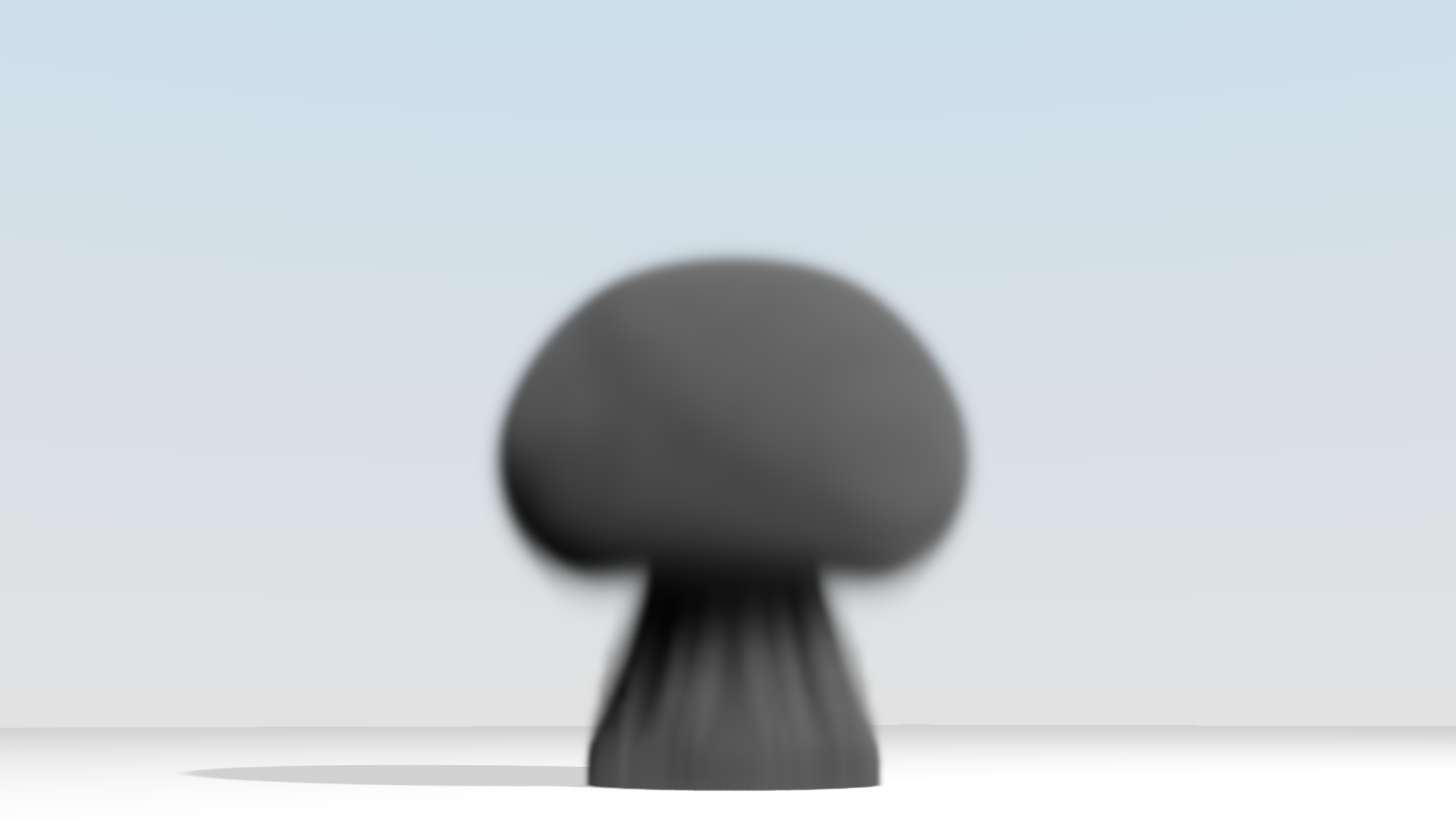}
}
\caption{Plume simulation (without vorticity confinement). \textit{Top left}: Jacobi (34 iterations). \textit{Top Middle} Jacobi (100 iterations). \textit{Top Right}: PCG. \textit{Bottom left}: Yang et al. \textit{Bottom middle}: small-model. \textit{Bottom Right}: this work.}
\label{fig:plumes}
\end{figure}

The supervised loss specified by Yang et al. measures the distance to a ground-truth output pressure (i.e. they train a network in isolation to perform the pressure projection only). For our dataset, this loss does not result in accurate results. When fluid cells are surrounded by solid cells, each connected component of fluid represents and independent linear system, each with an arbitrary pressure offset. As such, we modified the learning procedure of Yang et al. to include a ``pressure-normalization'' routine which subtracts the mean pressure in each connected-component of fluid cells in the ground-truth pressure frames. This modification enabled SGD to converge when training their model on our dataset.
However despite this correction, the model of Yang et al. does not learn an accurate linear projection on our data; our initial condition divergent velocity frames include large gradient and buoyancy terms, which results in a high amplitude, low frequency gradient in the ground-truth pressure. The small 3x3x3 input context of their model is not able to infer such low frequency outputs. Since these loss terms dominate the Yang et al. objective, the network over-trains to minimize it. Likely, this phenomena wasn't evident in Yang et al.'s original results due to the low diversity of their training set, and the high correlation between their evaluation and training conditions (perhaps their intended use-case). The results of their model trained on their objective function can be seen in Figure~\ref{fig:plumes}.

By contrast, our unsupervised objective minimizes divergence after the pressure gradient operator, whose FD calculation acts as a high-pass filter.
%, and as such the low frequency gravity and buoyancy terms do not dominate.
This is a significant advantage; our objective function is ``softer'' on the divergence contribution for phenomena that the network cannot easily infer. For the remaining experimental results, we will evaluate an improved version of the Yang et al. model, which we call our \emph{``small-model''}\footnote{The ``small-model'' is a single resolution pressure network with only 3x3x3 context and trained using the loss function, top-level architectural improvements and data-augmentation strategies from this work. We include these experiments as a proxy comparison for the original model of Yang et al.}.

\begin{figure}
\centerline{\includegraphics[width=0.85\columnwidth]{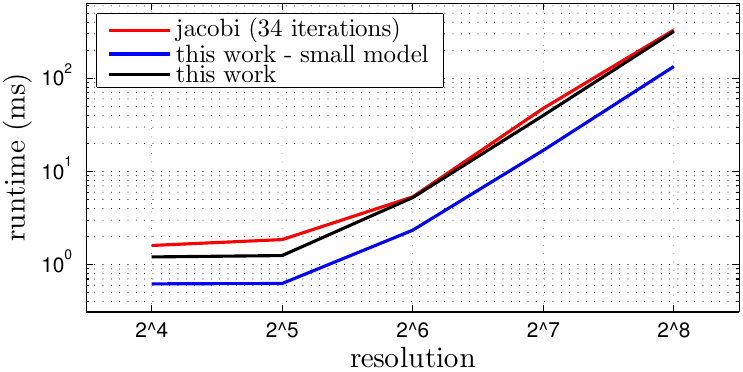}}
\caption{Pressure projection time (ms) versus resolution (PCG not shown for clarity).}
\label{fig:timing}
\end{figure}

Figure~\ref{fig:timing} shows the computation time of the Jacobi method, the small-model (sizing of Yang et al.) and this work\footnote{This runtime includes the pressure projection steps only: the velocity divergence calculation, the linear system solve and the velocity update. We use an NVIDIA Titan X GPU with 12GB of ram and an Intel Xeon E5-2690 CPU.}. Note that for fair quantitative comparison of output residual, we choose the number of Jacobi iterations (34) to approximately match the FPROP time of our network (i.e. to compare divergence with fixed compute). Since the asymptotic complexity as a function of resolution is the same for Jacobi and our ConvNet, the FPROP times are equivalent. PCG is orders of magnitude slower at all resolutions and has been omitted for clarity. The small-model provides a significant speedup over other methods. The runtime for the PCG, Jacobi, this work, and the small-model at $128^3$ grid resolution are 2521ms, 47.6ms, 39.9ms and 16.9ms respectively.

Note that with custom hardware~\cite{movidius,tpugoogle}, separable convolutions and other architectural enhancements, we believe the runtime of our ConvNet could be reduced significantly. However, we leave this to future work.

% Training takes approximate 20 and 48 hours for the 2D and 3D models respectively.

% An extremely important implementation detail to ensure proper convergence is that we clamp the L2 magnitude of the gradient to a constant value during training (we use an aggressive value of 1). The distribution of the gradient norm (L2) at the start of training has - unsurprisingly - high mean and low variance. Towards the end of training the mean gets close to zero, as training moves towards the local minima. However the variance actually increases significantly, and this can cause issues for the ADAM optimizer. We attribute high gradient variance to our semi-supervised long-term divergence term; when the ConvNet predicts frames with high divergence (as it has yet to finish training), for a small subset of training frames the divergence will grows significantly over the $n$ steps for our semi-supervised future divergence calculation. This can cause the gradient contributions for these frames to result in optimization instability. Clamping the gradient magnitude - thereby removing large gradient outliers - completely solves this issue and actually improves performance on the test-set.

\begin{figure}
\centering
\subfloat{%
\includegraphics[width=0.32\columnwidth,trim=14cm 0 12cm 0cm, clip=true]{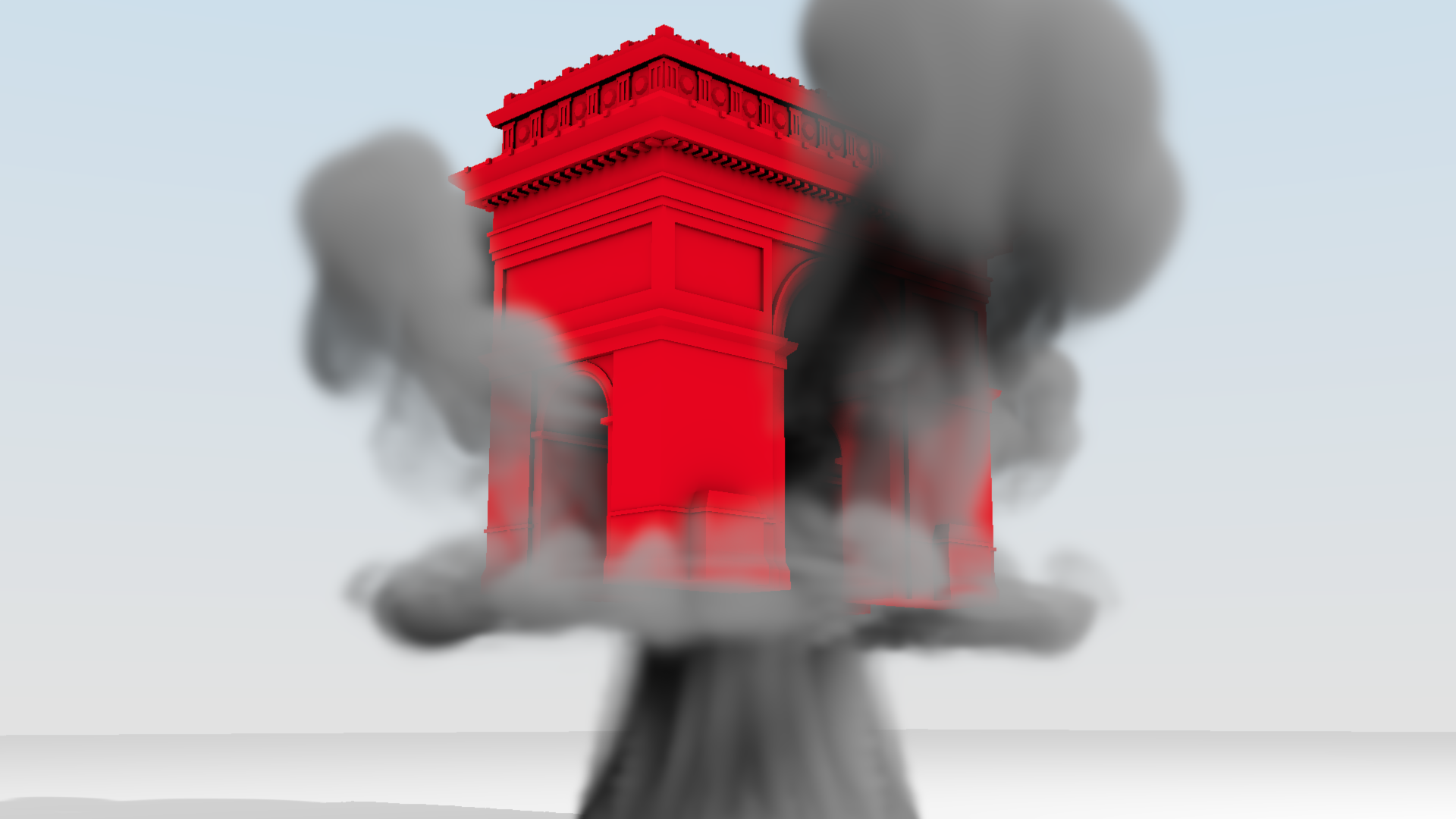}
}
\hfill
\subfloat{%
\includegraphics[width=0.32\columnwidth,trim=14cm 0 12cm 0cm, clip=true]{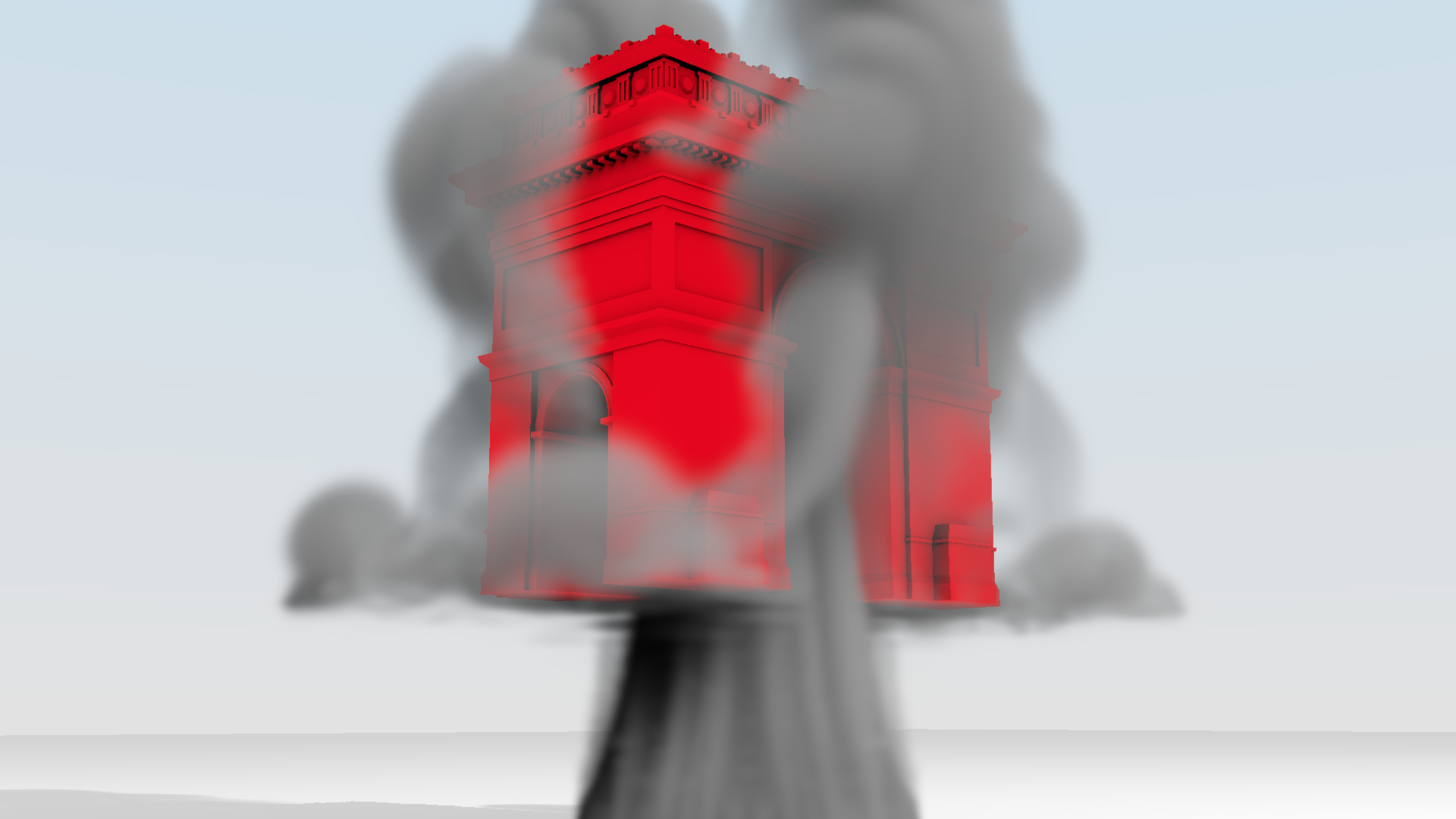}
}
\hfill
\subfloat{%
\includegraphics[width=0.32\columnwidth,trim=14cm 0 12cm 0cm, clip=true]{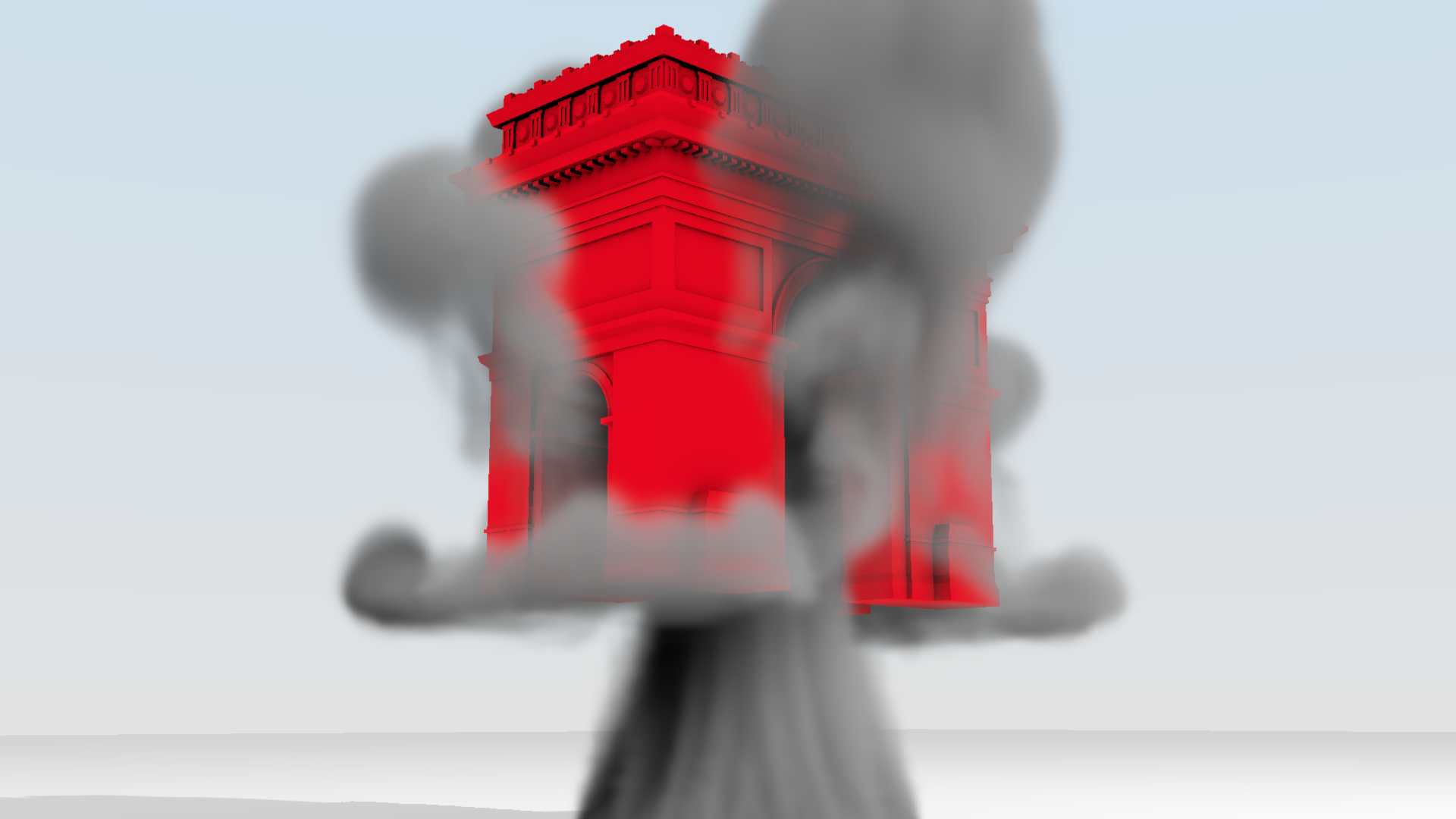}
}

\vspace{-10px}
\subfloat{%
\includegraphics[width=0.32\columnwidth,trim=22cm 3cm 30cm 22cm, clip=true]{figures/arch_pcg_frame_168.png}
}
\hfill
\subfloat{%
\includegraphics[width=0.32\columnwidth,trim=22cm 3cm 30cm 22cm, clip=true]{figures/arch_yang_frame_168.png}
}
\hfill
\subfloat{%
\includegraphics[width=0.32\columnwidth,trim=22cm 3cm 30cm 22cm, clip=true]{figures/arch_ours_frame_168.png}
}
\caption{Plume simulation with ``Arch'' geometry. \textit{Left}: PCG. \textit{Middle} small-model \textit{Right}: this work.}
\label{fig:arcplumes}
\end{figure}

We simulated a 3D smoke plume using both our system and baseline methods. The simulation data was created using our real-time system (which supports basic real-time visualization), and the accompanying figures and supplemental videos were rendered offline using Blender~\cite{blender}. Video examples of these experiments can be found in the supplemental materials.
Since vorticity confinement tends to obfuscate simulation errors (by adding high frequency detail), Figures~\ref{fig:plumes} and \ref{fig:arcplumes} were simulated without it.

Figure~\ref{fig:plumes} shows a rendered frame of our plume simulation (without geometry) for all methods at the same simulation time-step. Note that this boundary condition is not present in the training set and represents an input divergent flow 5 times wider than the largest impulse present during training. It is a difficult test of generalization performance.
Qualitatively, the PCG solver, 100 iteration Jacobi solver and our network produce visually similar results. The small-model's narrow receptive field cannot accurately simulate the large vortex under the plume, and as a result the plume rises too quickly (i.e. with too much upward velocity) and exhibits density blurring under the plume itself. The Jacobi method, when truncated early at 34 iterations, introduces implausible high frequency noise and has an elongated shape due to inaccurate modeling of buoyancy forces.
%
% Since our buoyancy, vorticity confinement and advection implementations differ significantly to those in mantaflow, quantitative comparison is not appropriate. However, qualitatively we do find that our system is more prone to early smoke dissipation than the full PCG solver and our system lacks some of the high-frequency smoke detail evident in mantaflow (the mantaflow smoke plume has both higher density and structural thickness). We attribute the increased numerical dissipation of our system to lack of model capacity and the network's subsequent (undesirable) tendency to increase dissipation over time in order to minimize long-term divergence. The trade-off of course is that the mantaflow PCG solver takes approximately 43 seconds per frame, whereas our system solver takes approximately 21ms per frame.
%
We also repeat the above simulation with solid cells from two models held out of our training set: the ``arc de triomphe'' model~\cite{3dmodels} and the Stanford bunny~\cite{stanfordbunny}; the single frame results for the arch simulation are shown in Figure~\ref{fig:arcplumes}. Since this scene exhibits lots of turbulent flow, qualitative comparison is less useful. The small-model has difficulty minimizing divergence around large flat boundaries and results in high-frequency density artifacts as shown. Both ConvNet based methods lose some smoke density inside the arch due to negative divergence at the fluid-geometry boundary (specifically at the large flat ceiling).

In addition to comparing divergence using fixed-compute, we also performed the above experiment by fixing divergence to $\operatorname{max}_t\left(||\nabla \cdot u_t||\right)$ = 0.872, and measuring compute. For Jacobi to match the divergence performance of our network, it requires 116 iterations and so is $4.1\times$ slower than our network. Since calculating divergence at inference time is fast and efficient, PCG can be used as a fallback method if our system fails with minimal runtime penalty or if the application at hand requires an exact solution.

Table~\ref{tab:norm_resid} shows the maximum norm L2 linear system residual for each frame over the entire simulation. The maximum residual for the Yang et al. model was greater than 1e5 for both simulations (i.e. it fails on this test case). Interestingly, early termination of the Jacobi method at 34 iterations results in reduced long-term accuracy and visual quality (Figure~\ref{fig:plumes}), however the maximum per-frame residual is still relatively low. \textit{This suggests that single frame divergence alone is not a sufficient condition to maintain long-term accuracy, and that the multi-frame error propagation mechanisms are an extremely important (but harder to quantify) factor.}
\begin{table}[ht]
\centering
\begin{footnotesize} % small, footnotesize, scriptsize, tiny
\begin{tabular}{ l c c c c }
  \hline
  \noalign{\vskip 1mm}
  
                                      & PCG   & Jacobi & small-model  & this work \\
  \noalign{\vskip 1mm}
  \hline
  \noalign{\vskip 1mm}
                     No geom            & $<$1e-3 & 2.44        & 3.436 & 2.482 \\
                     With geom  & $<$1e-3 & 1.235       & 1.966 & 0.872 \\               
  \noalign{\vskip 1mm}
  \hline
\end{tabular} 
\end{footnotesize}
\caption{Maximum residual norm throughout our Plume simulation with and without geometry ($\operatorname{max}_t\left(||\nabla \cdot u_t||\right)$}
\label{tab:norm_resid}
\end{table}
% For the experiments presented in this work, we use $\lambda=1$ for all $\lambda$ values in our objective function. However, adjusting $\lambda_{div}$ to 0.2 (or lower) can help reduce numerical dissipation at the cost of an increase in long-term divergence (and a subsequent softer constraint on conservation of mass). We believe this flexibility is a benefit of our machine learning based approach, where the optimization hyper-parameters can be chosen as per the user application requirements.

\begin{figure}
    \centerline{\includegraphics[width=0.9\columnwidth]{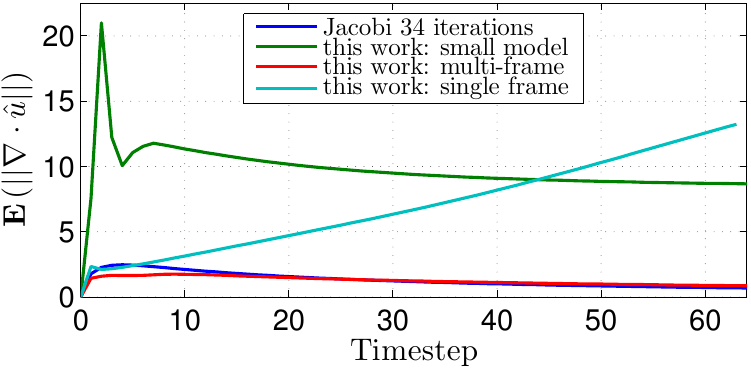}}
    \caption{$\mathbf{E}\left(\norm{\nabla \cdot \hat{u}_i}\right)$ versus time-step for each frame sample in our dataset (PCG not shown for clarity).}
    \label{fig:meanDivVsTime}
\end{figure}

% Of primary importance to real-time fluid simulation is measuring velocity field divergence over time to ensure stability and correctness. As such, we evaluate our model by seeding a simulation using the velocity field from manta, and by running the simulation for 256 frames; Figure~\ref{fig:divVsTime} shows $\norm{\nabla \cdot \hat{u}}$ for a random sub-set of the Manta Test-set initial conditions.
%
% Put figure here so it actually goes to the top of the correct page.
% \begin{figure}   \centerline{\includegraphics[width=\columnwidth]{figures/all_runs.pdf}}
%     \caption{$\norm{\nabla \cdot u_t}$ over time for all test-set initial conditions.}
%     \label{fig:divVsTime}
% \end{figure}
%
As a test of long-term stability, we record the mean L2 norm of velocity divergence ($\mathbf{E}\left(\norm{\nabla \cdot \hat{u}_i}\right)$) across all samples in our test-set. The result of this experiment is shown in Figure~\ref{fig:meanDivVsTime}. On our test-set frames, our method outperforms the small-model by a significant margin and is competitive with Jacobi truncated to 34 iterations. 
Figure~\ref{fig:meanDivVsTime} also shows the results of our model when a single time-step loss is used. Adding multi-frame components not only improves the divergence residual over multiple time steps as expected (since this is what we are directly minimizing), but additionally the single frame divergence performance is also improved. We attribute this to the fact that these future frames effectively increase dataset diversity, and can be seen as a form of dataset augmentation to improve generalization performance.
% \begin{figure}
%     \centerline{\includegraphics[width=\columnwidth]{figures/norm_div_longterm.pdf}}
%     \caption{$\mathbf{E}\left(\norm{\nabla \cdot \hat{u}_i}\right)$ versus time-step (frame count) with and without including long-term divergence minimization.}
%     \label{fig:meanDivVsTimeLongterm}
% \end{figure}

%<KRISTOFER CHANGE>
% REMOVED TO MAKE THE PAPER TAKE EXACTLY 8 PAGES
%One potential limitation with our approach is memory consumption. For a single frame inference, our 3D model uses 978kB of GPU RAM to store weights, biases and other network parameters (kept in GPU RAM to reduce PCI transfers), and requires a peak memory of 64.01MB to store the input and output tensors of the largest layer.
%</KRISTOFER CHANGE>

% **********************************************************************
\section{Conclusion}
% **********************************************************************
\label{sec:conclusion}

This work proposes a novel, fast and efficient method for calculating numerical solutions to the inviscid Euler Equations for fluid flow. 
We present a data-driven approach for approximate inference of the sparse linear system used to enforce the Navier-Stokes incompressibility condition - the ``pressure projection" step. We propose an unsupervised training-loss, which incorporates multi-frame information to improve long-term stability. We also present a novel and tailored ConvNet architecture, which facilitates drop-in replacement with existing Eulerian-based solvers.
While the proposed approach cannot guarantee finding an exact solution to the pressure projection step, it can empirically produce very stable divergence free velocity fields whose runtime and accuracy is better than the Jacobi method (a common technique used for real-time simulation) and whose visual results are comparable to PCG, while being orders of magnitude faster.
Code, data and videos are made available at
{\small\url{https://google.github.io/FluidNet}}.

%\appendix
%\section{Appendix Section}
%\label{sec:appendixsection}

% \section{Acknowledgements}
% We would like to thank the NVIDIA Corporation for their donation of a Tesla K40 used during this research, as well as K\'{a}roly Zsolnai-Feh\'{e}r for his invaluable advice and insights.

\bibliographystyle{icml2017}
\bibliography{main}

%\received{September 2008}{March 2009}

\end{document}